\documentclass[10pt,twocolumn,a4paper]{esaAI}

\title{Machine Learning-based vs Deep Learning-based Anomaly Detection in Multivariate Time Series for Spacecraft Attitude Sensors}

\def\authorEmail{riccardo.gallon@airbus.com}

\author[1, 2]{Riccardo Gallon\thanks{Riccardo Gallon. E-Mail: \authorEmail}}
\author[1]{Fabian Schiemenz}
\author[1]{Alisa Krstova}
\author[2]{Alessandra Menicucci}
\author[2]{Eberhard Gill}
\affil[1]{Airbus Defence and Space GmbH, Claude-Dornier Stra\ss e, Immenstaad am Bodensee,
88090, Germany.}
\affil[2]{Department of Space Systems Engineering, Faculty of Aerospace Engineering, Delft University of Technology, Kluyverweg 1, HS Delft 2629, Netherlands.}

\begin{document}

\makeCustomtitle

\begin{abstract}
In the framework of Failure Detection, Isolation and Recovery (FDIR) on spacecraft, new AI-based approaches are emerging in the state of the art to overcome the limitations commonly imposed by traditional threshold checking.

The present research aims at characterizing two different approaches to the problem of stuck values detection in multivariate time series coming from spacecraft attitude sensors. The analysis reveals the performance differences in the two approaches, while commenting on their interpretability and generalization to different scenarios.
\end{abstract}

\section{Introduction}

Failure Detection, Isolation, and Recovery (FDIR) onboard satellites is responsible of monitoring onboard parameters and functionalities, supervising the progression of the operations of the satellite under nominal conditions.

Newest trends in onboard FDIR \cite{mess2019techniques}, \cite{murphy2021machine}, \cite{tipaldi2015survey} leverage Artificial Intelligence (AI) to face the issues traditionally affecting the methods based on the ECSS-E-ST-70-41C (Packet Utilization Standard, PUS, \cite{ECSS-E-ST-70-41C}), currently representing the state of the art for flying missions. 
These issues mainly derive from the PUS-based FDIR relying on threshold- and expected values-monitoring, which make it inherently unable to detect certain kinds of data and/or anomalies, i.e. multivariate signals or univariate signals evolving anomalously in the threshold range.

AI-based solutions are expected to act on these limitations, aiming at enhancing detection notice and failures symptoms recognition.

\subsection{Project Framework} \label{subs:proj_fram}

The project \textit{Astrone KI}, Developed at Airbus Defence and Space GmbH in Friedrichshafen, Germany, is a collaborative effort with the Universities of Stuttgart and Dresden, and ASTOS Solutions GmbH. It  consists of a concept study for a drone-like vehicle for the exploration of Small Solar System Bodies, designed to perform autonomous relocation in an asteroid environment, aided by AI-augmented FDIR and vision-based navigation.

The \textit{Astrone KI} system is meant to operate the AI-based FDIR functionalities alongside the PUS-based FDIR in order to prove the effectiveness of an innovative technology on a critical task while relieving the criticality of its decision-making. Indeed, an effective AI integration in the PUS-based FDIR can make use of the AI-enhanced detection, while relying on the traditional FDIR as fallback in those cases where the AI could fail or is not required.

In \textit{Astrone KI}, the AI algorithms are meant to run on a dedicated AI module (i.e. a dedicated hardware), which not only performs the inference, but also the data preprocessing and postprocessing, including telemetry (TM) and telecommand (TC) handling.

The onboard sensors of the \textit{Astrone KI} spacecraft, which will be subject to AI-based FDIR, include accelerometers and Inertial Measurement Units (IMU), respectively measuring accelerations and angular rates in the shape of multivariate time series. 

The failure modes for the mentioned sensors are derived by Reliability, Availability, Maintainability and Safety (RAMS)  analysis, with special consideration to those failures which may benefit from the AI introduction. In the case of the multivariate time series coming from the accelerometer and the IMU of \textit{Astrone KI}, the analysis concentrated on stuck value faults. These faults constitute a suitable use-case for the AI-based FDIR functionalities, as in their presence the signal evolves anomalously within the ranges of thresholds monitored by the PUS-based FDIR. Classical methods can commonly detect stuck values only as long as the consequences propagate on higher levels (e.g. subsystem- or system-level), ultimately requiring a more drastic recovery. The AI introduction aims at detecting the stuck value directly at its occurrence in a particular signal, while being insured against missed detection by the presence of the classic PUS-based FDIR.
%
\subsection{Literature Review}

The literature research for the present work oriented towards AI-based anomaly detection for time series from both inside and outside the space domain, due to the generality of AI approaches in the field. In order to employ these algorithms onboard satellites, an important constraint is always represented by the limited computational resources, that pose limitations to the range of applicable solutions.

Commonly employed anomaly detection approaches make use of Machine Learning (ML) or Deep Learning (DL) to process onboard telemetry \cite{mess2019techniques}. ML approaches are based on classifying the data into nominal and faulty, employing deterministic algorithms capable of learning from data. Typical examples of this category include Support Vector Machines (SVM, \cite{xiong2011anomaly}) and Decision Trees \cite{kuhn2013decision}, \cite{costa2023recent}, which are classification algorithms employed in direct faults classification. More specifically, \textit{XGBoost} is a Decision Tree algorithm based on Gradient Boosting \cite{kuhn2013decision}, proven to beat state-of-the-art performances on Kaggle public datasets \cite{chen2016xgboost}, which has found large employment in the field of anomaly detection \cite{chalapathy2019deep}, \cite{braei2020anomaly}. 

DL solutions represent a class of black-box algorithms learning from the data to accomplish a specific task, detaching from the ML counterpart due to the lack of explainability of the input-output link, as well as the interpretability of the learnt features. DL is inherently able to grasp complex patterns in big datasets with limited or absent feature engineering, making it suitable to solve complex tasks where traditionally ML has proven its inefficiency (e.g. computer vision), or introducing solutions where there were none before (e.g. large language models). Recent scientific literature has seen a raising of DL due to the mentioned reasons. 

DL solutions can be used for direct fault classification of the input signal and for time series forecasting or reconstruction, in this case needing a further step for the actual classification. DL solutions (i.e. Neural Networks) are consistently employed in the field of anomaly detection, both for direct classification purposes \cite{mansell2021deep}, and for reconstruction \cite{hundman2018detecting}. An example of a Convolutional Neural Network (CNN) for direct anomaly classification in the industrial field is given in \cite{cui2016multi}. The same task applied to simulated spacecraft onboard telemetry can be found in \cite{mansell2021deep}, which employs a Long Short-term Memory (LSTM) cell with a preprocessing stage making use of an SVM. Finally, \cite{hundman2018detecting} is a pioneering work in the field of time series forecasting and classification in the space domain, proposing an approach of anomaly detection based on the comparison between the signal time series coming from the real unit and its forecast counterpart. 

\subsection{Contribution of the Present Work}

In the present work, two approaches to the problem of stuck values detection in time series are proposed and applied to the use-case of the \textit{Astrone KI} system.

On one side, the ML approach \textit{XGBoost} is employed, focusing on the interpretability of the algorithm, designing it to mimic human-based stuck values recognition through a specific selection of hyperparameters. On the other side, a CNN is proposed to accomplish the same detection task, proving its outstanding performances at the cost of losing any interpretability and generalization capability.

\section{Results} \label{sec:results}

\subsection{Experimental Setup}

\begin{figure}[h!]
    \centering
    \includegraphics[width=0.7\linewidth]{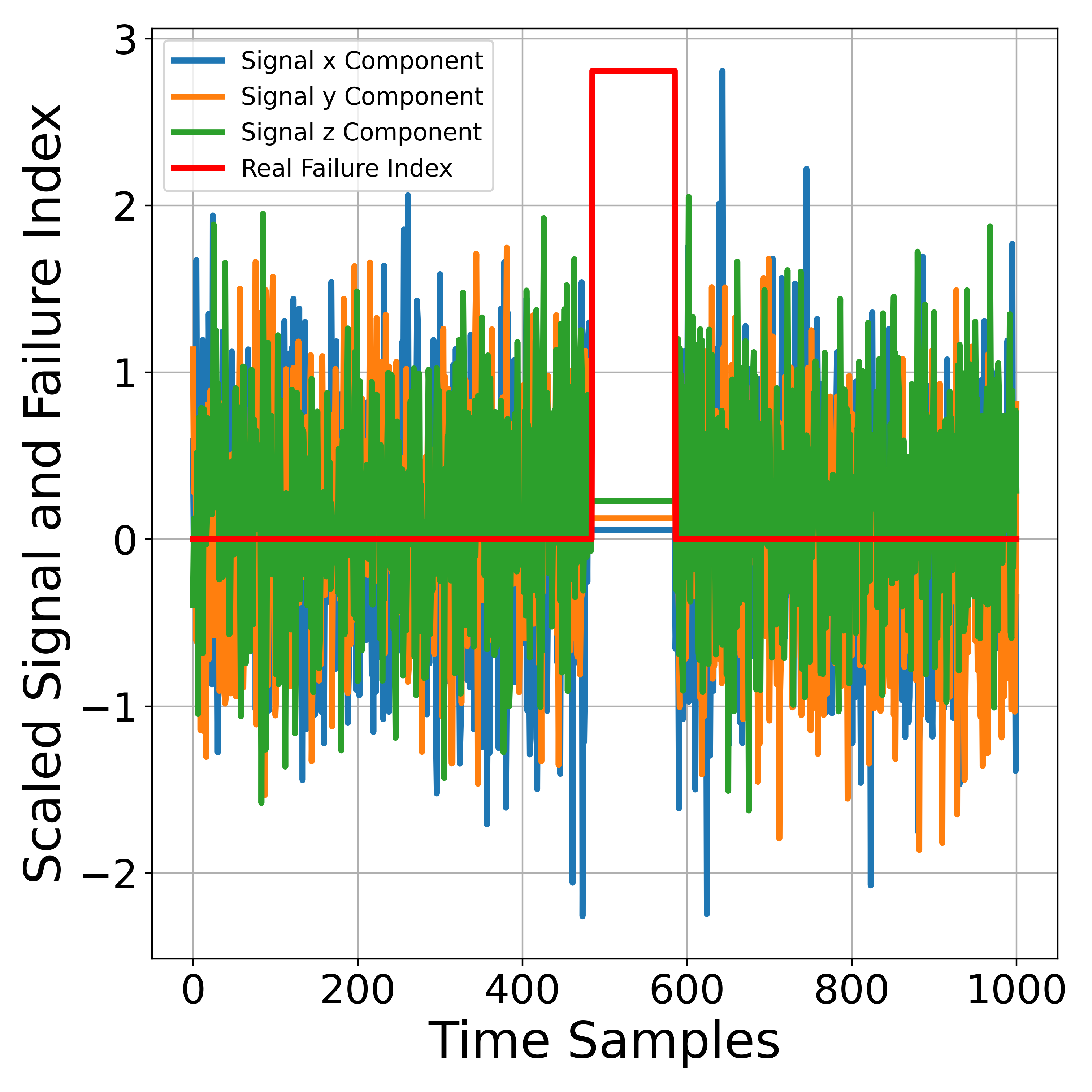}
    \caption{Example of a stuck at last value on all three axes of the accelerometer}
    \label{fig:ex_stuck_acc}
\end{figure}

\begin{figure}[h!]
    \centering
    \includegraphics[width=0.7\linewidth]{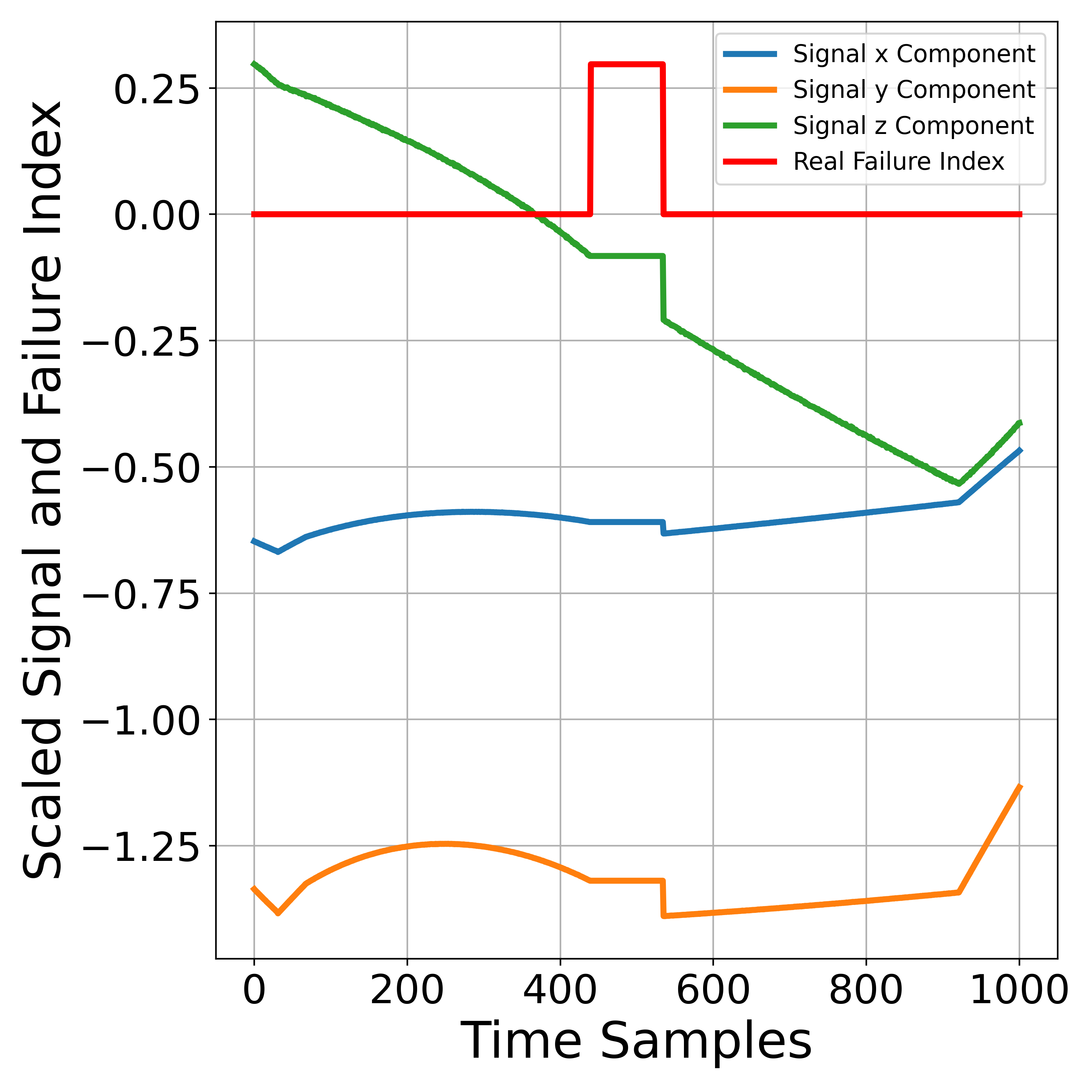}
    \caption{Example of a stuck at last value on all three axes of the IMU}
    \label{fig:ex_stuck_imu}
\end{figure}

As already discussed in \cref{subs:proj_fram}, the AI-based FDIR strategy is focused on detecting stuck values (\cref{fig:ex_stuck_acc} and \cref{fig:ex_stuck_imu}). These faults realize in various combinations of three defining parameters: the signal value, stuck at the last value, or stuck at random value; the axis of occurrence, whether it affects a single random axis or all three axes of the sensor simultaneously; and the presence or absence of measurement noise on top of the fault.

By means of the Attitude and Orbit Control System (AOCS) Offline Simulation Environment (AOSE, \cite{wiedermannmodel}), the unit models of the \textit{Astrone KI} sensors are simulated to generate the dataset for the AI training and testing. Each sensor is simulated over a set of spacecraft trajectories obtained from randomized initial conditions, and stuck values are randomly injected into the data. The criteria for the fault injection stage is meant to replicate realistic faults occurrence across the single trajectories, at the same time maintaining balance between nominal and faulty data. 

Key intuition of the present analysis is that stuck value failures are essentially slices of a signal where the derivative sharply drops to zero, eventually following a sharp peak (if stuck at random). The peak is directly related to the sudden change in the signal's value between two successive time samples, as its derivative is computed numerically. The approached zero value is instead influenced by measurement noise (if present), which induces zero-mean oscillations in the signal's derivative.

The conditions for identifying a stuck value are summarized in \cref{tab:conditions_stuck}.
\begin{table}[h]\renewcommand{\arraystretch}{1.2}
\centering
\begin{tabular}{p{0.5\linewidth} || c}
\hline\hline
Condition & Failure            \\
\hline\hline
signal value significantly outside the nominal range  & stuck at random      \\
Peak in signal derivative & stuck at random        \\
Zero signal derivative & stuck at random/last\\
\hline\hline
\end{tabular}
\caption{Conditions to Recognize Stuck Values}
\label{tab:conditions_stuck}
\end{table}

The approaches considered for the stuck values detection are based on ML and DL respectively.
The first solution presented employs \textit{XGBoost}. The algorithm has been designed to obtain a structure of the computed tree reflecting the rules expressed in \cref{tab:conditions_stuck}, pursuing full interpretability of the classification process. Also, \textit{XGBoost} has been employed to classify faults in the IMU signal only, as the reduced noise content of this sensor is expected to facilitate the application of the conditions in \cref{tab:conditions_stuck}.

The second approach presented is a Multi-channel Convolutional Neural Network \cite{cui2016multi}, which is capable of detecting stuck values in the accelerometer and IMU signals coupled. The Network takes as input a sliding window over the sensors signals, with a fixed length, and classifies each final sample of the window based on the preceding signal history as included in the window itself. 

The preference for using a CNN over other common approaches (i.e. RNN), is based on two main reasons. First, although RNNs are typically employed for time series due to their temporal analysis capabilities, CNNs can effectively leverage their spatial approach in the specific stuck value detection task presented. The chosen input features allow to focus on the signal's shape within individual windows to detect typical stuck value artifacts (i.e. flat signal). \\
Second, CNNs are significantly more parallelizable than RNNs and generally require fewer parameters to accomplish the same tasks. This results in lighter models and faster inference. Light computational resources and speed are particularly important, given the limited computational resources available in state-of-the-art space hardware, whether it be an OBC or dedicated hardware (e.g. FPGA) as required in \textit{Astrone KI}.

The output of both \textit{XGBoost} and the Neural Network is made by binary indexes marking the presence of a failure in the sensors separately, regardless of the occurring fault case or the interested axis. This choice reflect the consideration that the recovery action which is eventually triggered in the onboard FDIR would be the same for all the considered fault cases.

\subsection{Training}

The hyperparameters employed in \textit{XGBoost} training are: \textit{binary crossentropy} as loss function, specifically for the binary classification task; \textit{trees number} set to 1 and \textit{max depth} per tree set to 6. The number of trees refers to the number of classifiers trained at each step of the Gradient Boosting algorithm, while the max depth indicates the maximum number of \textit{splits} per tree. Motivation to this specific hyperparameter setup can be once again found in the rules outlined in \cref{tab:conditions_stuck} and further explained in \cref{sec:discussion}.

The CNN is made by two convolutional + max pooling stages, followed by a concatenation layer and a last convolution + max pooling layer. The \textit{binary crossentropy} is employed for the binary classification task, while the other hyperparameters, including the learning rate and the filters' size and number, are subject to optimization to maximize classification scores. The number of samples per input window is set to 180 (subject to optimization).

\cref{tab:metrics} presents a comparison of the performances obtained with the two presented algorithms.
\begin{table}[h]\renewcommand{\arraystretch}{1.2}
\centering
\begin{tabular}{c || c || c}
\hline\hline
\textbf{Model} & \textit{\textbf{Precision}} & \textit{\textbf{Recall}}            \\
\hline\hline
\textit{XGBoost}    &  0.98 & 0.75   \\
CNN                 &  0.97 & 0.96   \\
\hline\hline
\end{tabular}
\caption{\textit{XGBoost} vs CNN performances on a test set drawn from the IMU dataset}
\label{tab:metrics}
\end{table}

\section{Discussion} \label{sec:discussion}

The two anomaly detection approaches proposed in Section \ref{sec:results} present advantages and drawbacks, which are highlighted in the following analysis. 

The most favorable feature on the \textit{XGBoost} side is the clear understanding of the \textit{rules} followed to classify the input. This feature, especially when led by robust feature engineering, constitutes the main advantage of ML algorithms over their DL counterparts. However, since the features choice is always made by the algorithm developer, it may sometimes result in suboptimal performances with respect to the unknown problem optimum. Neural Networks suffer much less this issue, because good performances can be achieved also with limited or absent feature engineering. Nevertheless, concerning the specific onboard FDIR application of the present work, an interpretable algorithm is deemed more dependable with respect to a Neural Network for the mentioned interpretability feature, therefore it is considered more likely to be employed in flying missions.

The hyperparameters employed to train \textit{XGBoost} serve the purpose of making the algorithm reason as exposed in \cref{tab:conditions_stuck}. In other words, the algorithm is designed to learn \textit{splits} and \textit{rules} classifying data based on the derivative value, the presence of peaks, and the signal value going significantly out of range. An analysis of the tree structure after training (not reported here for brevity) proves the realization of this reasoning.

In conclusion, the task left to the algorithm is only to quantify the numerical thresholds required by the conditions in \cref{tab:conditions_stuck}.

As a further proof that the algorithm is reasoning as intended, also suggesting its generalization capabilities, an experiment of transfer learning was carried out by inferencing the \textit{XGBoost} model on the accelerometer signal (\cref{fig:ex_stuck_acc}). The experiment shows that the algorithm achieves comparable performances to the IMU case in the detection of stuck values with zero derivative (no noise, corresponding to a dead interface).
\begin{table}[h]\renewcommand{\arraystretch}{1.2}
\centering
\begin{tabular}{c || c || c}
\hline\hline
\textbf{Model} & \textit{\textbf{Precision}} & \textit{\textbf{Recall}}            \\
\hline\hline
\textit{XGBoost}    &  0.99 & 0.68   \\
CNN                 &  0.99 & 0.96   \\
\hline\hline
\end{tabular}
\caption{\textit{XGBoost} vs CNN performances on a test set drawn from the accelerometer dataset}
\label{tab:metrics_acc}
\end{table}

A deeper analysis of the fault predictions shows that all the false negatives (undetected faults) correspond to stuck values with noise, highlighting the biggest inherent limitation of the \textit{XGBoost} approach. The measurement noise is indeed responsible of leading the derivative value beyond the thresholds posed by the Decision Tree in the learnt \textit{splits}, determining a systematic error that it is not possible to delete by only acting on the algorithm. Note that this effect implies a higher deterioration of the \textit{recall} in the accelerometer case with respect to the IMU, given the more noisy nature of the signal.

The CNN approach generally outperforms \textit{XGBoost} in terms of performance metrics (\cref{tab:metrics} and \cref{tab:metrics_acc}), while losing all interpretability. Additionally, it has the capability to simultaneously analyze both signals, potentially gathering valuable information for detection from both sources. Furthermore, unlike \textit{XGBoost}, the CNN exhibits no systematic limitations in its classification capability, theoretically granting the same classification ability across all considered failure cases. Nevertheless, results show that the CNN struggles with classifying part of the stuck values with noise, which primarily contribute to the deterioration of \textit{recall} (\cref{tab:metrics} and \cref{tab:metrics_acc}). This confirms the complexity of classifying that specific fault case, as well as suggesting that the CNN is improving the \textit{XGBoost} performances while reasoning similarly.

The cost for this CNN performance improvement is obviously the interpretability, as the Neural Network does not offer a clear insight on its way of reasoning as the decision tree does, preventing among the others any possible consideration on its limitations. Traditionally, Neural Networks can achieve outstanding performances, but need to be carefully maintained to ensure they keep performing correctly. Even a slight shift in the input data distribution with respect to the training data while operating the network may  make a fine-tuning and/or retrain necessary \cite{huyen2022designing}.

As a final remark, when it comes to choose between the CNN and \textit{XGBoost} in a real scenario, it is important to keep into account case-dependent requirements, e.g. favor interpretability over performances, or maximize case-specific metrics not mentioned in this study. In this regard, the development of these metrics considering the real-time nature of the application, or also the impact of the AI on the system level, can be an interesting field for future work.

\section*{Acknowledgements}

The results presented in this paper have been achieved by the project Astrone - Increasing the Mobility of Small Body Probes, which has received funding from the German Federal Ministry for Economic Affairs and Energy (BMWi) under funding number “50 RA 2130A”. The consortium consists of Airbus Defence and Space GmbH, Astos Solutions GmbH, Institute of Automation (Technische Universität Dresden) and Institute of Flight Mechanics and Controls (Universität Stuttgart). Responsibility of the publication contents is with the publishing author.

\printbibliography

@article{murphy2021machine,
  title={Machine Learning in Space: A Review of Machine Learning Algorithms and Hardware for Space Applications.},
  author={Murphy, James and Ward, John E and Mac Namee, Brian},
  journal={AICS},
  pages={72--83},
  year={2021}
}

@inproceedings{mess2019techniques,
  title={Techniques of artificial intelligence for space applications-a survey},
  author={Me{\ss}, Jan-Gerd and Dannemann, Frank and Greif, Fabian},
  booktitle={European workshop on on-board data processing (OBDP2019)},
  year={2019},
  organization={European Space Agency}
}

@article{tipaldi2015survey,
  title={Survey on fault detection, isolation, and recovery strategies in the space domain},
  author={Tipaldi, Massimo and Bruenjes, Bernhard},
  journal={Journal of Aerospace Information Systems},
  volume={12},
  number={2},
  pages={235--256},
  year={2015},
  publisher={American Institute of Aeronautics and Astronautics}
}

@techreport{ECSS-E-ST-70-41C,
    author = {ECSS-E-ST-70-41C},
    title = {Telemetry and telecommand packet utilization},
    institution = {European Space Agency},
    year = {2016},
    number = {ECSS-E-ST-70-41C},
}

@conference{wiedermannmodel,
  title        = {Model based design of the Sentinel-2 attitude control system},
  author={Wiedermann, G and Winkler, S and Gockel, W and Reggio, D and Levenhagen, J},
  year         = 2009,
  booktitle    = {Deutscher Luft- und Raumfahrtkongress},
  organization = {Deutsches Zentrum für Luft- und Raumfahrt }
}

@inproceedings{xiong2011anomaly,
  title={Anomaly detection of spacecraft based on least squares support vector machine},
  author={Xiong, Long and Ma, Hao-Dong and Fang, Hong-Zheng and Zou, Ke-Xu and Yi, Da-Wei},
  booktitle={2011 Prognostics and System Health Managment Confernece},
  pages={1--6},
  year={2011},
  organization={IEEE}
}

@article{costa2023recent,
  title={Recent advances in decision trees: An updated survey},
  author={Costa, Vin{\'\i}cius G and Pedreira, Carlos E},
  journal={Artificial Intelligence Review},
  volume={56},
  number={5},
  pages={4765--4800},
  year={2023},
  publisher={Springer}
}

@inbook{kuhn2013decision,
  title = {Applied Predictive Modeling},  
  author = {Kuhn, Max and Johnson, Kjell},
  year    = {2013},
  booktitle = {Applied Predictive Modeling},  
  publisher = {Springer},
  pages   = {369-413}
}

@article{cui2016multi,
  title={Multi-scale convolutional neural networks for time series classification},
  author={Cui, Zhicheng and Chen, Wenlin and Chen, Yixin},
  journal={arXiv preprint arXiv:1603.06995},
  year={2016}
}

@inproceedings{hundman2018detecting,
  title={Detecting spacecraft anomalies using lstms and nonparametric dynamic thresholding},
  author={Hundman, Kyle and Constantinou, Valentino and Laporte, Christopher and Colwell, Ian and Soderstrom, Tom},
  booktitle={Proceedings of the 24th ACM SIGKDD international conference on knowledge discovery \& data mining},
  pages={387--395},
  year={2018}
}

@article{mansell2021deep,
  title={Deep learning fault diagnosis for spacecraft attitude determination and control},
  author={Mansell, Justin R and Spencer, David A},
  journal={Journal of Aerospace Information Systems},
  volume={18},
  number={3},
  pages={102--115},
  year={2021},
  publisher={American Institute of Aeronautics and Astronautics}
}

@inproceedings{chen2016xgboost,
  title={Xgboost: A scalable tree boosting system},
  author={Chen, Tianqi and Guestrin, Carlos},
  booktitle={Proceedings of the 22nd acm sigkdd international conference on knowledge discovery and data mining},
  pages={785--794},
  year={2016}
}

@article{braei2020anomaly,
  title={Anomaly detection in univariate time-series: A survey on the state-of-the-art},
  author={Braei, Mohammad and Wagner, Sebastian},
  journal={arXiv preprint arXiv:2004.00433},
  year={2020}
}

@article{chalapathy2019deep,
  title={Deep learning for anomaly detection: A survey},
  author={Chalapathy, Raghavendra and Chawla, Sanjay},
  journal={arXiv preprint arXiv:1901.03407},
  year={2019}
}

@inbook{huyen2022designing,
  title = {Designing machine learning systems},
  author = {Huyen, Chip},
  year         = {2022},
  booktitle    = {Designing machine learning systems},
  publisher    = {O'Reilly Media, Inc.},
  pages        = {191--292}
}
\addcontentsline{toc}{section}{References}

\end{document}